\documentclass[11pt,a4paper]{article}
\usepackage[hyperref]{acl2021}
\usepackage{times}
\usepackage{latexsym}

\usepackage{microtype}

\usepackage{graphicx} 
\usepackage{amsmath}
\usepackage{bm}
\usepackage{threeparttable}
\usepackage{booktabs}
\usepackage{multirow}

\aclfinalcopy 

\title{{LICHEE}: Improving Language Model Pre-training \\with Multi-grained Tokenization}

\author{Weidong Guo$^{1}$\thanks{$^*$ Equal contribution.} , Mingjun Zhao$^{2}$\footnotemark[1] , Lusheng Zhang$^{1}$\footnotemark[1] , Di Niu$^{2}$, Jinwen Luo$^{1}$, \\
\textbf{Zhenhua Liu$^{1}$, Zhenyang Li$^{1}$, Jianbo Tang$^{1}$} \\
  $^{1}$Platform and Content Group, Tencent \\
  $^{2}$University of Alberta \\
  \tt \{weidongguo, lshzhang, jamsluo, edinliu,
  \\ \tt nickzyli, jianbotang\}@tencent.com, \\
  \tt \{zhao2, dniu\}@ualberta.ca
}

\date{}

\begin{document}
\maketitle

\begin{abstract}
Language model pre-training based on large corpora has achieved tremendous success in terms of constructing enriched contextual representations and has led to significant performance gains on a diverse range of Natural Language Understanding (NLU) tasks. 
Despite the success, most current pre-trained language models, such as BERT, are trained based on single-grained tokenization, usually with fine-grained characters or sub-words, making it hard for them to learn the precise meaning of coarse-grained words and phrases.
In this paper, we propose a simple yet effective pre-training method named \textit{LICHEE} to efficiently incorporate multi-grained information of input text. Our method can be applied to various pre-trained language models and improve their representation capability. Extensive experiments conducted on CLUE and SuperGLUE demonstrate that our method achieves comprehensive improvements on a wide variety of NLU tasks in both Chinese and English with little extra inference cost incurred, and that our best ensemble model achieves the state-of-the-art performance on CLUE benchmark competition.

\end{abstract}

\section{Introduction}


Pre-trained language models (PLMs) such as GPT \citep{radford2018improving}, BERT \citep{devlin2019bert} and XLNet \citep{yang2019xlnet} have become enormously popular and achieved great success on diverse natural language understanding tasks, such as sentiment analysis, question answering, and language inference. These models usually utilize a transformer architecture \cite{vaswani2017attention} to capture the dependencies between tokens in the input text, to model the language information, and to learn  contextual representations. It is first pre-trainined based on large-scale unlabeled corpora, and subsequently fine-tuned based on the labeled data from downstream tasks.

In many NLU applications, tokenization often affects the performance and needs to be chosen carefully.
The input tokens for pre-trained language models are usually fine-grained, e.g., words and sub-words for English and characters for Chinese.
Compared with coarse-grained tokens such as phrases, 
the advantage of fine-grained tokens is that they form a smaller vocabulary, yielding abundant training samples per token, and thus alleviating the data sparsity issue and out-of-vocabulary (OOV) problem \citep{li2019word}. However, even trained on large corpora, it is still hard for language models pre-trained with fine-grained tokens to learn the correct attention boundaries of larger semantic units in many languages \citep{zhang2020ambert}.

To obtain a more accurate model, prior studies attempt to incorporate coarse-grained information into models trained with fine-grained tokenization by masking sequences of consecutive tokens in the pre-training stage \citep{joshi2020spanbert,cui2019pre}.
\citet{zhang2020ambert} propose AMBERT, a Siamese network based on BERT to handle multi-grained input text, and uses two encoders with shared weights to separately encode fine-grained tokens and coarse-grained tokens into two sequences of contextualized representations. Despite its effectiveness, the inference cost of AMBERT almost doubles that of the original BERT due to the dual-encoder structure, which is often unacceptable in industrial scenarios.

In this paper, we propose a novel method named \textit{LICHEE} designed to efficiently leverage the input information at multiple levels of granularity in the pre-training stage in order to enhance the representation ability of PLMs.  
Unlike AMBERT that encodes the fine-grained and coarse-grained tokens with two encoders, which significantly increases the inference cost, in \textit{LICHEE} the fusion of multi-grained information of input text happens at the embedding level, which requires no change on the original model structure of the PLM, and thus induces little extra inference cost when applied in online NLP applications.
Specifically, \textit{LICHEE} first pre-processes the input text into fine-grained and coarse-grained tokens, which are passed through two embedding layers, respectively, to derive their corresponding vector representations. Both vector representations are then merged via pooling to form the \textit{multi-grained embedding vector}, which serves as the input to the PLM encoder. Finally, the enhanced contextual representations generated by the PLM encoder, with both fine-grained and coarse-grained information incorporated, are obtained and used for downstream tasks.

We have applied \textit{LICHEE} to enhance multiple different pre-trained language models, including BERT \citep{devlin2019bert}, ALBERT \citep{lan2019albert}, and GPT \citep{brown2020language}, and conducted extensive evaluation of the resulted language models on Chinese natural language understanding (NLU) tasks evaluated by CLUE \citep{CLUEbenchmark} benchmarks. 
Results show that with \textit{LICHEE}, the resulted pre-trained language models significantly outperform their single-grained counterparts on almost all tasks, by taking advantage of multi-grained information to effectively and efficiently produce more accurate representations.

In addition, we also participated in the CLUE benchmark competition with our best ensemble model built upon a collection of \textit{LICHEE}-enhanced BERT-large models, and achieved the state-of-the-art performance of an average score of $80.42$ (as of January 8, 2021) over 9 different Chinese NLU tasks, as well as the best scores on two individual tasks: IFLYTEK and CSL.

Moreover, we have also conducted English natural language understanding experiments based on SuperGLUE \citep{wang2019superglue} benchmarks. Significant improvements are observed when \textit{LICHEE} is employed in the pre-training stage, which demonstrates that the proposed pre-training method is generally effective in different language settings.

\section{Related Work}

In this section, we give a brief overview of some popular pre-trained language models and studies on the training techniques related to tokenization.

Pre-trained language models are pre-trained on large unsupervised corpora and aim to produce meaningful representations for each input token not only considering the meaning of itself, but also with its surrounding contexts anticipated.
ELMo \citep{peters2018deep} is one of the first pre-trained language models based on bidirectional LSTMs which produces the contextual representation of each token by concatenating its left-to-right and right-to-left representations.
GPTs \citep{radford2018improving,radford2019language,brown2020language} leverage the powerful Transformer \citep{vaswani2017attention} to build an auto-regressive language model predicting the next token given its history context.
BERT \citep{devlin2019bert} is a bidirectional auto-encoding language model also based on transformer. 
It consists of two pre-training objectives: masked language model (MLM) and next sentence prediction (NSP).
\citet{yang2019xlnet} point out the discrepancy of the pre-training and fine-tuning stage of BERT due to the masking symbol, and propose a permutation language model called XLNet \citep{yang2019xlnet}.

The great popularity of BERT draws many researchers to make improvements on the architecture.
RoBERTa \citep{liu2019roberta} improves several training details of BERT including dynamic masking and  the removal of the NSP pre-training task.
ALBERT \citep{lan2019albert} reduces the model parameters with cross-layer weight sharing and accelerates the training process.
ELECTRA \citep{clark2019electra} proposes a new token detection task and adopts a  generator-discriminator framework to pre-train the language model.

Although most pre-trained language models are built on fine-grained tokenization, coarse-grained information proves to be helpful to the model performance.
\citet{cui2019pre} propose a masking scheme called ``whole word masking'' (WWM) for Chinese BERT, where the consecutive characters belonging to the same word are masked together.
In ERNIE \citep{sun2019ernie}, knowledge graphs are added to enhance the model, and entity level masking is used during the pre-training, which is beneficial for language understanding tasks. 
SpanBERT \citep{joshi2020spanbert} proposes to mask random spans instead of random tokens, and adopts a new span boundary objective task to replace the next sentence prediction task in the pre-training.
Instead of focusing on the masking scheme, AMBERT \citep{zhang2020ambert} proposes to adopt two encoders with shared parameters to learn the representations of fine-grained and coarse-grained tokens in parallel. 
However, even that the weight sharing setting reduces the number of model parameters, the dual-encoder structure of AMBERT induces twice the inference cost, which remains a huge issue when deployed in online applications.

Different from AMBERT, our work merges the fine-grained and coarse-grained tokenization at embedding level, and achieves significant performance gains with little additional computation costs.

\begin{figure*}[tb]
	\centering
	\includegraphics[width=\textwidth]{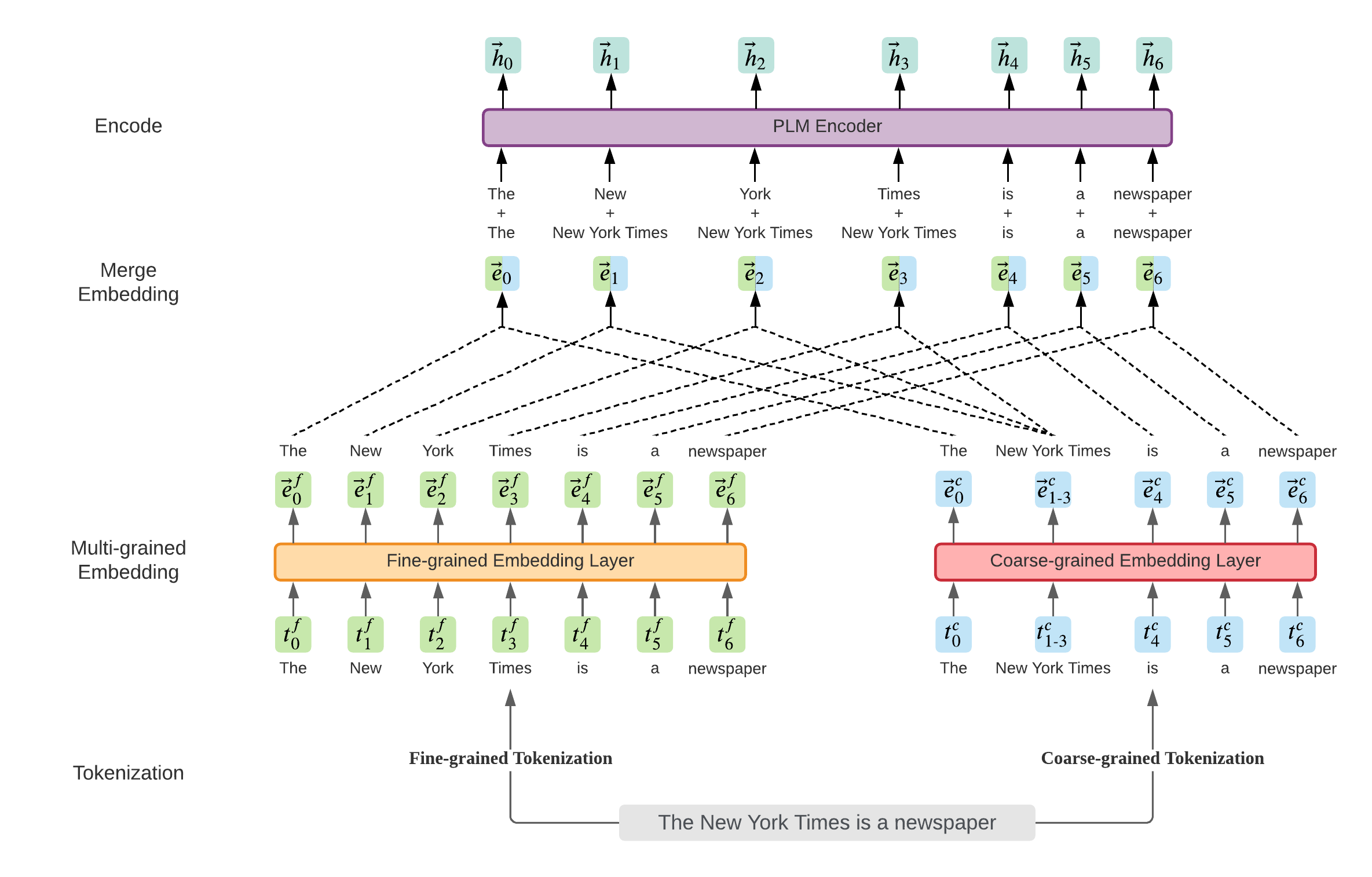}
	\caption{The overall structure of our proposed pre-training framework \textit{LICHEE}. Fine-grained and coarse-grained tokens are first derived from the input text by tokenization, and separately passed into two individual embedding layers. The multi-grained embedding vectors are acquired by taking a max-pooling on the fine-grained and coarse-grained embedding vectors, and are fed into the PLM encoder to extract the final contextualized representations.}
	\label{fig:Framework}
\end{figure*}

\section{Methodology}


In this section, we present \textit{LICHEE}, the general multi-grained framework for language model pre-training, and its detailed implementation, including the pre-training methods for both auto-regressive and auto-encoding tasks and fine-tuning details.

\subsection{Model Architecture}
Figure \ref{fig:Framework} gives an overview of \textit{LICHEE} where the input information from multiple granularities is leveraged to enhance the representation ability for many pre-trained language models.

The framework takes in text sequences as input which are tokenized into token sequences.
In this paper, we keep two vocabularies and use two tokenizers to perform fine-grained and coarse-grained tokenizations, where items in vocabularies are selected based on their token frequencies in pre-training corpora.
Also, the definitions of ``fine grain'' and ``coarse grain'' vary across languages.
For example, in English, words and phrases are often used as the fine-grained and coarse-grained tokens respectively. And in Chinese, characters and words are used instead.
Officially, for a given input text sequence $T$, we use $t_i^f$ to denote the $i$-th fine-grained token and $t_{j\textit{-}k}^c$ to denote a coarse-grained token that is composed of fine-grained tokens $\{t_j^f,...,t_k^f\}$ between $j$ and $k$.
For example, in figure \ref{fig:Framework}, the coarse-grained token ``New York Times'' is composed of the first, sencond, and third fine-grained tokens, and is denoted as $t_{1\textit{-}3}^c$.


After tokenization, two separate embedding layers are used to map the tokenized tokens to their vector representations. 
Specifically, each fine-grained token $t_i^f$ is passed into a fine-grained embedding layer to produce the fine-grained embedding vector $\vec{e}_i^f \in \mathcal{R}^{d}$ of the token, where $d$ denotes the dimension of the fine-grained embedding.
Similarly, the coarse-grained embedding $\vec{e}_{j\textit{-}k}^c \in \mathcal{R}^{d}$ is derived with the same dimension $d$ by feeding token $t_{j\textit{-}k}^c$ to the coarse-grained embedding layer, shown as:
\begin{equation}
    \begin{aligned}
        \vec{e}_i^f &= embedding_{fine}(t_i^f), \\
        \vec{e}_{j\textit{-}k}^c &= embedding_{coarse}(t_{j\textit{-}k}^c).
    \end{aligned}    
\end{equation}

For each token $t_i^f$, we construct its multi-grained embedding vector $\vec{e}_i \in \mathcal{R}^{d}$ by performing a max-pooling operation on the derived fine-grained embedding $\vec{e}_i^f$ and the coarse-grained embedding $\vec{e}_{j\textit{-}k}^c$ of its corresponding coarse-grained token $t_{j\textit{-}k}^c$:
\begin{equation}
    \begin{aligned}
        \vec{e}_i &= \textit{max-pool}(\vec{e}_i^f, \vec{e}_{j\textit{-}k}^c), 
    \end{aligned}    
\end{equation}
where $j \leq i \leq k$.
Note that $d$ is equal to the original embedding dimension of the single-grained PLM, to prove that the performance gain is contributed to the introduction of multi-grained information other than modified model structure.

Finally, the combined embedding vectors $\bm{\vec{e}}$ are fed into the PLM encoder to construct the final contextualized representations $\bm{\vec{h}}$ enhanced with multi-grained information:
\begin{equation}
    \begin{aligned}
        \bm{\vec{h}} = encode(\bm{\vec{e}}).
    \end{aligned}    
\end{equation}

\subsection{Pre-training}
We have applied \textit{LICHEE} on both auto-regressive and auto-encoding PLMs, such as GPT and BERT.

For \textbf{\textit{auto-regressive PLMs}}, the pre-training task is Next Token Prediction which aims to predict the next token $t_i$ based on its previous context $t_{<i}$, by optimizing the following objective function
\begin{equation}
    \begin{aligned}
        \min_\theta -\sum_i \log p_{\theta}(t_i|t_{<i}),
    \end{aligned}    
\end{equation}
where the conditional probability $p_\theta$ is modeled with a network with parameter $\theta$.

In our framework, we adjust the objective function to include both fine-grained context $t_{<i}^f$ and coarse-grained context $t_{<i}^c$, shown as:
\begin{equation}
    \begin{aligned}
        \min_\theta -\sum_i \log p_{\theta}(t_i|t_{<i}^f, t_{<i}^c).
    \end{aligned}    
\end{equation}
Note that when making predictions on any token within a coarse-grained span $t_i \in t_{j\textit{-}k}^c$, the token embedding $\vec{e}_i$ will cause information leakage as it involves the coarse-grained token embedding $\vec{e}_{j\textit{-}k}^c$ which contains information beyond the history context.
For example, in the case illustrated in figure \ref{fig:Framework}, the prediction on token ``York'' should not rely on token ``New'' and its embedding $\vec{e}_1$ as it discloses the entire information of the coarse-grained token of ``New York Times'' by the coarse-grained embedding $\vec{e}_{1\textit{-}3}^c$.
Therefore, we can only exploit the context before the start position of the coarse-grained token to make predictions, illustrated as:
\begin{equation}
    \begin{aligned}
        \min_\theta -\sum_{j \leq i \leq k} \log p_{\theta}(t_i|t_{<j}^f, t_{<j}^c),
    \end{aligned}    
\end{equation}
where $j$ and $k$ are the start and end positions of the coarse-grained token.

For \textbf{\textit{auto-encoding PLMs}}, we only include Masked Language Modeling (MLM) task in the pre-training process, as Next Sentence Prediction (NSP) task is shown to have no benefits indicated in many recent studies \cite{lan2019albert,liu2019roberta,zhang2020ambert}.
In MLM, 15\% of the tokens are randomly selected and substituted with a set of tokens, in which 80\% are replaced with \textit{[MASK]} token, 10\% are replaced with random tokens, and 10\% stay unchanged.

The objective is to recover the masked tokens $T^m \subset T$ from the altered text input sequence $\tilde{T}$:
\begin{equation}
    \begin{aligned}
        \min_\theta -\sum_{t^m \in T^m} \log p_{\theta}(t^m|\tilde{T}).
    \end{aligned}    
\end{equation}

In our framework, we propose to exploit the multi-grained information of the input in the MLM task, shown as:
\begin{equation}
    \begin{aligned}
        \min_\theta -\sum_{t^m \in T^m} \log p_{\theta}(t^m|\tilde{T}^f, \tilde{T}^c),
    \end{aligned}    
\end{equation}
where $\tilde{T}^f$ and $\tilde{T}^c$ stand for the fine-grained and coarse-grained altered input text.

Similar to the strategy deployed in auto-regressive PLMs, we apply a masking strategy that when a fine-grained token $t_i^f$ is to be masked, its corresponding coarse-grained token $t_{j\textit{-}k}^c$ and all the fine-grained tokens $t_j^f,...,t_k^f$ belonging to it are also masked, in order to avoid information leakage from the multi-grained embeddings.

\subsection{Fine-tuning}
In fine-tuning of downstream tasks, we append the special tokens (\textit{[CLS]}, \textit{[SEP]}) to both fine-grained and coarse-grained vocabularies.
In sentence-level classification tasks, \textit{[CLS]} is attached to the start of input sequences in auto-encoding PLMs like BERT, and to the end of the input in auto-regressive PLMs like GPT. Its multi-grained contextualized representation $\vec{h}_{[CLS]}$ is used to represent the whole input sequence and is passed into a projecting layer for the final prediction.

Similarly, for tasks that include token-level span detection, such as Question Answering, the contextual representation $\vec{h}_i$ for each token $t_i$ is extracted and utilized in the task.

\section{Experiments}

We have carried out extensive experiments on various natural language understanding tasks on both Chinese and English datasets.
In the following section, we will first introduce the pre-training datasets used in our evaluation and provide the implementation details of our framework.
And we demonstrate the effectiveness of \textit{LICHEE} by conducting comprehensive experiments on various Chinese NLU datasets with multiple different PLMs, and compare our method with other baseline methods.
Next, we perform a thorough ablation study to evaluate different approaches of integrating input text information from multiple granularities.
Finally, we adopt \textit{LICHEE} to an English BERT to verify its efficacy on English NLU tasks.

\subsection{Pre-Training Datasets}
For Chinese language, there is no commonly used corpus for pre-training language models. 
We utilize a large corpus consisting of $450$G text from a wide range of popular Chinese applications including Kandian, Zhihu, Wechat, and Weibo, in various fields of news, wiki, and blogs.

Similar to most Chinese PLMs, characters are used as fine-grained tokens due to the language nature of Chinese.
For coarse-grained tokens, We use QQSeg which is a segmentation tool with an open API to perform segmentation on text, and the segmented words are treated as coarse-grained tokens.
For the construction of vocabularies, we follow Google's Chinese BERT and include $21,128$ tokens in the fine-grained vocabulary.
And in the coarse-grained vocabulary, we calculate the token frequencies and trimmed out tokens with frequency lower than $8$, resulting in $210,946$ tokens.
Note that in order to alleviate the out-of-vocabulary (OOV) problem, all tokens in the fine-tuned vocabulary are also included in coarse-grained vocabulary.

For English, a corpus with $6.2$ million documents ($18.9$G compressed text) from Wikipedia is leveraged to pre-train the model.
We first perform sub-word tokenization with BPE algorithm \citep{sennrich2015neural} on the English text, where the produced words and sub-words constitute the fine-grained vocabulary of $28,996$ tokens.
In the coarse-grained vocabulary, we treat high-frequency words as coarse-grained tokens, resulting in $136,630$ tokens in total, which also include all tokens in the fine-grained vocabulary for the OOV concern.

\begin{table*}[tb]
  \resizebox{\textwidth}{!}{
  \begin{tabular}{l|c|ccccccccc}
    \toprule
    \multirow{2}{*}{\textbf{Model}} & \textbf{Avg.} & \textbf{TNEWS} & \textbf{IFLYTEK} & \textbf{AFQMC} & \textbf{OCNLI} & \textbf{CLUEWSC} & \textbf{CSL} & \textbf{CMRC2018} & \textbf{CHID} & \textbf{C3} \\
     & - & acc. & acc. & acc. & acc. & acc. & acc. & EM. & acc. & acc.\\
    \midrule
	BERT & $71.12$ & $66.62$ & $60.64$ & $71.74$ & $73.45$ & $72.92$ & $84.01$ & $73.08$ & $75.52$ & $62.08$ \\
	BERT-\textit{LICHEE} & $\mathbf{73.92}$ & $\mathbf{67.94}$ & $\mathbf{60.94}$ & $\mathbf{73.65}$ & $\mathbf{75.85}$ & $\mathbf{81.03}$ & $\mathbf{84.51}$ & $\mathbf{75.84}$ & $\mathbf{77.65}$ & $\mathbf{67.84}$ \\
    \midrule
    ALBERT & $67.27$ & $64.45$ & $57.54$ & $\mathbf{71.35}$ & $69.19$ & $68.80$ & $83.00$ & $68.06$ & $68.97$ & $54.04$ \\
    ALBERT-\textit{LICHEE} & $\mathbf{69.30}$ & $\mathbf{66.31}$ & $\mathbf{58.29}$ & $70.95$ & $\mathbf{71.05}$ & $\mathbf{70.39}$ & $\mathbf{83.31}$ & $\mathbf{72.87}$ & $\mathbf{71.93}$ & $\mathbf{58.65}$ \\
    \midrule
	GPT & $67.41$ & $67.52$ & $60.84$ & $69.83$ & $70.91$ & $63.76$ & $83.12$ & $62.53$ & $73.31$ & $54.84$ \\
	GPT-\textit{LICHEE} & $\mathbf{68.73}$ & $\mathbf{68.40}$ & $\mathbf{61.06}$ & $\mathbf{70.00}$ & $\mathbf{72.01}$ & $\mathbf{66.01}$ & $\mathbf{83.23}$ & $\mathbf{64.57}$ & $\mathbf{74.02}$ & $\mathbf{59.27}$ \\
    \bottomrule
  \end{tabular}
  }
  \caption{Comparison of the model performances on the CLUE tasks. BERT-\textit{LICHEE}, ALBERT-\textit{LICHEE} and GPT-\textit{LICHEE} stand for the multi-grained version of the model with our method incorporated. The average score of the nine CLUE tasks are also given.}
  \label{tab:clue-compare}
\end{table*}

\begin{table*}[tb]
  \resizebox{\textwidth}{!}{
  \begin{tabular}{l|c|ccccccccc}
    \toprule
    \multirow{2}{*}{\textbf{Model}} & \textbf{Avg.} & \textbf{TNEWS} & \textbf{IFLYTEK} & \textbf{AFQMC} & \textbf{OCNLI} & \textbf{CLUEWSC} & \textbf{CSL} & \textbf{CMRC2018} & \textbf{CHID} & \textbf{C3} \\
     & - & acc. & acc. & acc. & acc. & acc. & acc. & EM. & acc. & acc.\\
    \midrule
    Archer-24E-SINGLE & $79.19$ & $69.54$ & $62.27$ & $\mathbf{77.26}$ & $\mathbf{83.57}$ & $90.00$ & $85.73$ & $75.65$ & $85.66$ & $\mathbf{83.04}$ \\
	roberta\_selfrun & $79.46$ & $69.10$ & $63.92$ & $76.09$ & $80.40$ & $\mathbf{93.10}$ & $87.27$ & $79.20$ & $88.80$ & $77.29$ \\
	UER-ensemble & $79.64$ & $\mathbf{72.20}$ & $64.00$ & $76.82$ & $80.80$ & $90.35$ & $85.83$ & $79.15$ & $86.03$ & $81.60$ \\
	BERTs & $79.66$ & $69.94$ & $63.92$ & $76.77$ & $82.09$ & $88.97$ & $86.77$ & $\mathbf{80.50}$ & $\mathbf{89.51}$ & $78.44$ \\
	\midrule
	\textit{LICHEE}-ensemble & $\mathbf{80.06}$ & $70.50$ & $\mathbf{64.15}$ & $76.98$ & $81.30$ & $90.69$ & $\mathbf{87.40}$ & $79.80$ & $87.51$ & $82.22$ \\
    \bottomrule
  \end{tabular}
  }
  \caption{Top-5 models on the CLUE benchmark leaderboard where our ensemble model achieves the state-of-the-art performance on the averaged CLUE score. These results are grabbed from the official CLUE website\footnotemark on Jan 8, 2021.}
  \label{tab:clue-benchmark}
\end{table*}

\subsection{Benchmarks}
The evaluation of the pre-trained models is conducted on various downstream NLU tasks.
In our experiments, all the Chinese PLMs are evaluated on Chinese Language Understanding Evaluation (CLUE) \citep{CLUEbenchmark} which is a comprehensive language understanding benchmark developed for Chinese containing 9 natural language understanding tasks.
Within the 9 tasks, there are two single-sentence classification tasks that are TNEWS and IFLYTEK, four sentence-pair classification tasks that are AFQMC, OCNLI, CLUEWSC and CSL, and three question answering tasks that are CMRC2018, CHID, and C3.
Note that OCNLI has replaced CMNLI since Oct 22, 2020.
We compare the model performance by reporting the performance score of each task and the average score of all tasks.

For English tasks, we use the SuperGLUE benchmarks \citep{wang2019superglue} which is an extension of GLUE \citep{wang2019glue} consisting of a collection of 8 NLU tasks of higher difficulty for comprehensively evaluating the performance of English PLMs.
SuperGLEU contains a word sense disambiguation task (WiC), two textual entailment tasks (CB and RTE), two reasoning tasks (COPA and WSC), and three question answering tasks (BoolQ, MultiRC, and ReCoRD).

\subsection{Experiment Setup}
In order to demonstrate the general applicability and effectiveness of our framework, we have implemented three different pre-trained language models with our method including BERT, ALBERT and GPT, and compare the performances with their corresponding single-grained baseline methods.

For BERT and ALBERT, we follow the ``base'' structure in \citep{devlin2019bert} with an encoder of 12 layers.
And the GPT model in our experiment is also made up of a $12$-layer transformer decoder.
Then, we apply the following training setting to the training process of all three models.
For better scalability in large batch, we adopt LAMB \citep{you2019large} to replace Adam \citep{kingma2014adam} as the optimizer with a batch size of $768$ and a learning rate of $2e-4$.
We first train the model for $1$M steps using $128$ as the maximum sequence length, and increase the maximum length to $512$ for another $100$k steps, for better capturing the long distance dependencies.
To enhance the training efficiency, we adopt mix-precision training technique \citep{micikevicius2017mixed} during pre-training, which are performed on 4 Nvidia V100 gpus.

We have also implemented a \textit{LICHEE}-enhanced ensemble model based on BERT-large to participate in the CLUE benchmark competition.
During training, we adapt the batch size to $1,024$ and the maximum sequence lengths at the first and second stage are set to $256$ and $512$. 
And $64$ Nvidia V100 gpus are used to train the model.

For the evaluation of each task, we derive $6$ results with different random seeds and report the average performance in this paper.

\footnotetext{https://www.cluebenchmarks.com/rank.html}

\begin{table*}[tb]
  \resizebox{\textwidth}{!}{
  \begin{tabular}{l|c|ccccccccc}
    \toprule
    \multirow{2}{*}{\textbf{Model (BERT)}} & \textbf{Avg.} & \textbf{TNEWS} & \textbf{IFLYTEK} & \textbf{AFQMC} & \textbf{OCNLI} & \textbf{CLUEWSC} & \textbf{CSL} & \textbf{CMRC2018} & \textbf{CHID} & \textbf{C3} \\
     & - & acc. & acc. & acc. & acc. & acc. & acc. & EM. & acc. & acc.\\
    \midrule
	SG & $71.12$ & $66.62$ & $60.64$ & $71.74$ & $73.45$ & $72.92$ & $84.01$ & $73.08$ & $75.52$ & $62.08$ \\
	SG (WWM) & $72.24$ & $66.87$ & $60.55$ & $72.62$ & $74.41$ & $74.07$ & $84.12$ & $75.22$ & $77.74$ & $64.60$ \\
	MG (CAT 384+384) & $72.86$ & $\mathbf{68.11}$ & $61.09$ & $72.33$ & $75.08$ & $75.26$ & $84.48$ & $75.35$ & $77.84$ & $66.17$ \\
	MG (CAT 256+512) & $72.94$ & $67.63$ & $\mathbf{61.55}$ & $71.96$ & $74.97$ & $76.54$ & $84.16$ & $75.31$ & $78.17$ & $66.15$ \\
	MG (CAT 512+256) & $73.08$ & $67.88$ & $61.06$ & $73.07$ & $75.84$ & $74.45$ & $\mathbf{84.74}$ & $74.44$ & $\mathbf{78.29}$ & $\mathbf{67.91}$ \\
	MG (MEAN) & $73.22$ & $67.85$ & $60.99$ & $73.44$ & $\mathbf{75.97}$ & $76.31$ & $84.52$ & $75.54$ & $77.84$ & $66.53$ \\
	\textit{LICHEE} & $\mathbf{73.92}$ & $67.94$ & $60.94$ & $\mathbf{73.65}$ & $75.85$ & $\mathbf{81.03}$ & $84.51$ & $\mathbf{75.84}$ & $77.65$ & $67.84$ \\
    \bottomrule
  \end{tabular}
  }
  \caption{Ablation study of different pre-training strategies with BERT model on CLUE dataset. Two single-grained (SG) baselines and five multi-grained (MG) methods (\textit{LICHEE} and its variants) with different ways of integrating the fine-grained and coarse-grained representations are evaluated.}
  \label{tab:clue-ablation}
\end{table*}

\subsection{Main Results}
In table \ref{tab:clue-compare}, we adopt our multi-grained pre-training method on three pre-trained language models: BERT, ALBERT, and GPT, and compare them with their single-grained baselines on CLUE benchmark.
From the results, we can see that our method achieves significant performance gains by exploiting the multi-grained information of the text input.
The averaged CLUE scores of our multi-grained BERT-\textit{LICHEE}, ALBERT-\textit{LICHEE} and GPT-\textit{LICHEE} are $73.92$, $69.30$ and $68.73$ respectively, producing significant absolute improvements of $2.80$, $2.03$, and $1.32$ compared to their single-grained baseline models.
Aside from the improvement on the averaged CLUE score, it is also worth to mention that our multi-grained BERT-\textit{LICHEE} and GPT-\textit{LICHEE} outperforms their single-grained baselines on all $9$ NLU tasks in CLUE, while the ALBERT-\textit{LICHEE} model also beat the single-grained ALBERT in $8$ out of $9$ tasks, which provides strong evidence that the benefits of our method are generally applicable to different pre-trained language models and diverse NLU tasks.

In order to further investigate the potential of \textit{LICHEE}, we apply it on an ensemble model based on BERT-large and participate in the CLUE benchmark competition. 
As demonstrated in table \ref{tab:clue-benchmark}, our method outperforms all other candidates on the average score of $9$ CLUE tasks by a significant margin, and also achieves the state-of-the-art performance on two individual NLU tasks of IFLYTEK and CSL.
This results further proves that our multi-grained pre-training method is able to bring significant improvements on the representation ability of language models and is generally effective to a wide range of downstream NLU tasks.

The reason of \textit{LICHEE}'s success is that we adopt a multi-grained pre-training strategy to model the contextual information of the input text to leverage the advantages from both granularities, where fine-grained token representations are easier to learn considering the sufficient training samples, and coarse-grained tokens are more complete as lexical units and provide more accurate contextual information.
Furthermore, in our framework, the combination of the multi-grained information is realized on the embedding level so that we can keep the model structure unaltered, showing that the benefits are achieved entirely through the information gains caused by multi-grained pre-training other than model-level modifications.

\begin{table*}[tb]
  \resizebox{\textwidth}{!}{
  \begin{tabular}{l|c|cccccccc}
    \toprule
    \multirow{2}{*}{\textbf{Model}} & \textbf{Avg.} & \textbf{BoolQ} & \textbf{CB} & \textbf{COPA} & \textbf{MultiRC} & \textbf{ReCoRD} & \textbf{RTE} & \textbf{WiC} & \textbf{WSC}\\
     & - & acc. & acc. & acc. & EM. & EM. & acc. & acc. & acc.\\
    \midrule
	BERT-WWM & $63.64$ & $77.13$ & $79.76$ & $62.83$ & $24.88$ & $65.20$ & $70.88$ & $64.50$ & $63.94$ \\
    BERT-\textit{LICHEE} & $\mathbf{65.53}$ & $\mathbf{77.98}$ & $\mathbf{88.21}$ & $\mathbf{63.00}$ & $\mathbf{25.41}$ & $\mathbf{67.50}$ & $\mathbf{71.91}$ & $\mathbf{65.45}$ & $\mathbf{64.81}$ \\
    \bottomrule
  \end{tabular}
  }
  \caption{Comparison between our multi-grained BERT-\textit{LICHEE} and the single-grained BERT-WWM on SuperGLUE tasks.}
  \label{tab:glue-result}
\end{table*}

\begin{table}[tb]
  \centering
  \begin{tabular}{c|cc}
    \toprule
    Model & FLOPs & Speedup \\
    \midrule
    BERT & $43.5$B & $1.0$x \\
    AMBERT & $87.0$B & $0.5$x \\
    \textit{LICHEE} & $43.5$B & $1.0$x \\
    \bottomrule
  \end{tabular}
  \caption{Comparison of FLOPs and speedup among the single-grained BERT, AMBERT, and our method.}
  \label{tab:inference}
\end{table}

\subsection{Ablation Analysis}
We have conducted ablation analysis on CLUE benchmarks with BERT, to evaluate the impact of our multi-grained design, as well as perform a comprehensive study on the different methods of integrating the multi-grained embedding.
Table \ref{tab:clue-ablation} lists the performance of model variants with different training strategies, including two single-grained methods and five multi-grained methods.

The original single-grained BERT whose masking scheme is solely based on fine-grained tokens gives an average CLUE score of $71.12$.
The Whole Word Masking (WWM) technique \citep{cui2019pre} performs masking operations on continuous fine-grained tokens that form a coarse-grained token and improves the performance to $72.24$.
Note that although WWM utilizes coarse-grained token boundary information during the masking operations, it does not explicitly train representations for coarse-grained tokens.
Therefore, we treat WWM also as a single-grained pre-training method.

For multi-grained pre-training methods, we have conducted experiments to explore five different approaches of combining embedding representations of fine-grained and coarse-grained tokens, including concatenating the embedding vectors with different dimension settings, and integrating them with mean-pooling and max-pooling.
For the concatenation approaches, we keep the dimension of the concatenated multi-grained embeddings to $768$ to align with the baseline models, and apply three settings to adjust the dimensions of fine-grained and coarse-grained embedding correspondingly to ($384$, $384$), ($256$, $512$) and ($512$, $256$).
Empirically, we discover that the three concatenation settings achieve similar performances, while having larger embedding vectors for fine-grained tokens and smaller embedding vectors for coarse-grained tokens produces a slightly better performance of $73.08$ average CLUE score.

Exploiting mean-pooling to integrate the multi-grained information gives more performance gains compared with concatenation methods and reaches $73.22$ average CLUE score, which may be attributed to the greater number of embedding parameters, as pooling methods do not require a shrink on the embedding dimension and allow both fine-grained and coarse-grained embedding dimension to stay $768$.
Finally, \textit{LICHEE} with the max-pooling incorporated outperforms all the fore-mentioned approaches, attains an overall score of $73.92$, and achieves the best score on 3 out of 9 CLUE tasks, due to its capability of extracting more representative features.
Especially for the task of CLUEWSC, \textit{LICHEE} acquires an accuracy of $81.03$ while the second best method only reaches $76.54$.
We believe this is because the small training set of CLUEWSC with only $532$ examples makes it more dependent on powerful pre-trained representations, so that the advantage of the max-pooling method is amplified.

Overall, we can see from table \ref{tab:clue-ablation} that all multi-grained pre-training methods outperform the single-grained baselines by a significant margin, which again proves that our idea of incorporating multi-grained information during the pre-training phase is efficacious and can benefit model performance considerably.

\subsection{Inference Speed Analysis}
We have also studied the inference speed of \textit{LICHEE} and compare it with the original single-grained BERT and another multi-grained method AMBERT.

Table \ref{tab:inference} gives a brief comparison in terms of FLOPs and speedup, tested on a binary classification task with $512$ sequence length.
FLOPs indicates the number of floating-point operations that the model performs for a single process, where generally speaking, the higher the model's FLOPs is, the slower the inference speed will be. 

We can see that the FLOPs of the AMBERT is $87.0$ billion, twice the number of the single-grained BERT.
It means the inference time of AMBERT is almost doubled, which can cost a lot more time and resources, and often can be unacceptable for real-world applications.
Meanwhile, our multi-grained method produces a model with $43.5$ billion FLOPs with a negligible increase compared with the single-grained baseline, because the additional operations only include an embedding lookup operation for coarse-grained tokens and a max-pooling operation to integrate the fine-grained and coarse-grained embedding vectors.
In summary, \textit{LICHEE} can produce significant performance gains with negligible extra inference time needed.

\subsection{English Tasks}
We have also conducted experiments on SuperGLUE benchmarks to evaluate \textit{LICHEE} on English language tasks, and compared it with the single-grained baseline: BERT-WWM \citep{cui2019pre}.

As shown in table \ref{tab:glue-result}, the BERT model pre-trained with our multi-grained method outperforms the single-grained BERT-WWM on all $8$ SuperGLUE tasks, and attains an average score of $65.53$ surpassing the baseline by $1.89$.
This improvement over BERT-WWM demonstrates that the effectiveness of \textit{LICHEE} is attributed greatly to the information gain of its multi-grained representations, more than just token boundary information. 
We also notice that, similar to the CLUEWSC task, a huge increase of $8.45$ on accuracy is achieved for the CB dataset of $250$ training samples, because our pre-training method leverages the information gains of multi-grained tokens and produces more accurate representations, which is especially effective on tasks with small training data.

This result evidently illustrates that \textit{LICHEE} is not only effective on tasks of character based language like Chinese that highly relies on correct tokenizations, but can also produce significant improvements on languages that are naturally tokenized such as English.


\section{Conclusion}
In this paper, we have proposed a novel multi-grained method for language model pre-training named \textit{LICHEE}, which can be applied to both auto-regressive and auto-encoding PLMs.
In our method, the fine-grained embeddings and the coarse-grained embeddings are separately learned and integrated as the multi-grained embeddings, which is then passed into the encoder of the language model.
Experiments show that \textit{LICHEE} can significantly enhance the model performance by a great margin on downstream tasks of both Chinese and English, and significantly improve the inference speed compared to the prior multi-grained method.

\bibliography{main}
\bibliographystyle{acl_natbib}

\end{document}